\documentclass{article}

 \usepackage[preprint]{neurips_2025}

\usepackage[utf8]{inputenc} 
\usepackage[T1]{fontenc}    
\usepackage{hyperref}       
\usepackage{url}            
\usepackage{booktabs}       
\usepackage{amsfonts}       
\usepackage{nicefrac}       
\usepackage{microtype}      
\usepackage{xcolor}         
\usepackage{amsmath}
\usepackage{wrapfig}
\usepackage{subcaption}

\usepackage{multirow}
\usepackage{graphicx}
\usepackage[ruled,vlined]{algorithm2e}

\newcommand{\ourmlong}{Multi-level Aware Preference Learning}
\newcommand{\ourm}{MAPL}

\title{Multi-Level Aware Preference Learning: Enhancing RLHF for Complex Multi-Instruction Tasks}

\author{%
  Ruopei Sun \And Jianfeng Cai \And Jinhua Zhu \AND Kangwen Zhao \And Dongyun Xue
  \And Wengang Zhou \And Li Li \And Houqiang Li \\
  \AND \textnormal{University of Science and Technology of China}\\
  \texttt{ruopeisun@mail.ustc.edu.cn} \\
}

\begin{document}

\maketitle

\begin{abstract}
    Reinforcement Learning from Human Feedback (RLHF) has emerged as a predominant approach for aligning artificial intelligence systems with human preferences, demonstrating exceptional and measurable efficacy in instruction following tasks; however, it exhibits insufficient compliance capabilities when confronted with complex multi-instruction tasks. Conventional approaches rely heavily on human annotation or more sophisticated large language models, thereby introducing substantial resource expenditure or potential bias concerns. Meanwhile, alternative synthetic methods that augment standard preference datasets often compromise the model's semantic quality. Our research identifies a critical oversight in existing techniques, which predominantly focus on comparing responses while neglecting valuable latent signals embedded within prompt inputs, and which only focus on preference disparities at the intra-sample level, while neglecting to account for the inter-sample level preference differentials that exist among preference data. To leverage these previously neglected indicators, we propose a novel Multi-level Aware Preference Learning (MAPL) framework, capable of enhancing multi-instruction capabilities. Specifically, for any given response in original preference data pairs, we construct varied prompts with a preference relation under different conditions, in order to learn intra-sample level preference disparities. Furthermore, for any given original preference pair, we synthesize multi-instruction preference pairs to capture preference discrepancies at the inter-sample level. Building on the two datasets constructed above, we consequently devise two sophisticated training objective functions. Subsequently, our framework integrates seamlessly into both Reward Modeling and Direct Preference Optimization paradigms. Through rigorous evaluation across multiple benchmarks, we empirically validate the efficacy of our framework.
\end{abstract}

\section{Introduction}
\label{sec:1}

Large Language Models (LLMs)~\citep{yang2024qwen2technicalreport, bai2023qwentechnicalreport, grattafiori2024llama3herdmodels, achiam2023gpt, team2024gemini, liu2024deepseek, deepseekai2025deepseekr1incentivizingreasoningcapability, touvron2023llama2openfoundation} powered by Reinforcement Learning from Human Feedback (RLHF)~\citep{ziegler2019fine, stiennon2020learning, ouyang2022training} have achieved remarkable efficacy across diverse tasks~\citep{bai2022training, poesia2022synchromesh, 10.1145/3672456, hao-etal-2023-reasoning, zhao2024large, imani2023mathprompter, luo2023wizardmath}. RLHF has emerged as a powerful technique for aligning intelligent agents with human preferences by training a reward model to represent these preferences, which can then be used to train other models through reinforcement learning. Many works enhance the original RLHF method, further augmenting instruction-following capabilities of models~\citep{mukobi2023superhfsupervisediterativelearning, dong2024selfplayexecutionfeedbackimproving, stolfo2025improvinginstructionfollowinglanguagemodels, zhang2023wisdomhindsightmakeslanguage}.
However, although recent LLMs demonstrate relatively good performance on simple instructions, their response quality to complex instructions with multiple constraints often falls below expectations, with certain constraints being omitted or wrongly followed, which hinders their application in more realistic complex scenarios~\citep{he2024largelanguagemodelsunderstand}.

To address the aforementioned issues, previous research can be primarily categorized into three approaches:
1) Human annotation-based methods~\citep{DatabricksBlog2023DollyV2, köpf2023openassistantconversationsdemocratizing}. 
These methods enhance model instruction following capabilities by manually constructing datasets from scratch through human annotations.
2) Multi-instruction data augmentation of existing datasets using large language models~\citep{xu2023wizardlmempoweringlargelanguage, hui2024smallerlanguagemodelsbetter, sun2024coniferimprovingcomplexconstrained, huang2025muscimprovingcomplexinstruction, he2024complexsimpleenhancingmulticonstraint, zhang2024iopoempoweringllmscomplex}.
These methods augment existing datasets by using large language models to generate multi-instruction queries or to produce responses to multi-instruction queries.
3) Data augmentation of existing datasets through programmatic construction of multi-instruction data~\citep{yuan2024followinglengthconstraintsinstructions}. This approach is applied to existing preference datasets by programmatically constructing additional instructions to generate multi-instruction preference datasets. 

Nevertheless, these conventional approaches demonstrate notable limitations. They either rely heavily on human annotations or advanced large language models, introducing substantial resource requirements and potential subjective biases, or implement synthetic augmentation methods that often compromise semantic coherence. Our analysis identifies critical oversights in existing approaches, which disproportionately emphasize response comparison while failing to exploit the rich latent signals inherently embedded in prompt inputs, and which concentrate solely on intra-sample level preference distinctions while neglecting the inter-sample preference distinctions that present among preference data.

Based on the aforementioned observations, to effectively leverage these previously overlooked indicators, we propose a novel \ourmlong{} (\ourm{}) framework, which mainly comprises the following components:
1) {\bf Intra-sample level preference learning.} For any given response $y$ in original preference data, we construct varied prompts $x_w$ and $x_l$ with a preference relation under different conditions, where $x_w$ is better than $x_l$, to form a intra-sample level preference pair $(x_w, x_l, y)$. We design a sophisticated training objective function to guide the model in learning the intra-sample level preference disparities within these preference data.
2)~{\bf Inter-sample level preference learning.} For any given original preference pair $(x, y_w, y_l)$ where $x$ is the prompt, $y_w$ is the response considered better by humans and $y_l$ is the response considered worse by humans, we synthesize multi-instruction preference pairs $(x', y_w, y_l)$, where the concatenation of $x$ with a multi-instruction constraint forms a new prompt denoted as $x'$. These new pairs $(x', y_w, y_l)$ establishes preference data pairs at the inter-sample level in conjunction with the original preference pairs $(x, y_w, y_l)$. We then formulate another sophisticated training objective function that guides the model to discern inter-sample level preference disparities among these preference data.

To demonstrate the efficacy of our approach, we conduct comprehensive experiments on diverse models, including LLaMA3~\citep{grattafiori2024llama3herdmodels} and Qwen2~\citep{yang2024qwen2technicalreport}, and evaluate the proposed framework using several benchmarks. Compared to alternative methods, models trained by using our framework exhibit significant improvements in multi-instruction following capabilities without semantic degradation.

The main contributions of the paper are summarized as follows: (1) We propose a novel \ourmlong{} (\ourm{}) framework that enhances model capabilities in following multiple instructions concurrently. (2) We demonstrate that our approach can be seamlessly integrated into reward modeling and policy optimization frameworks with minimal architectural modifications. (3) Experimental results conclusively demonstrate the superiority of our framework in multi-instruction following performance compared to existing approaches.
\section{Related Work}

\paragraph{Reinforcement Learning from Human Feedback} Reinforcement Learning from Human Feedback~(RLHF)~\citep{ouyang2022traininglanguagemodelsfollow} represents a dominant paradigm for human preference alignment, involving reward model~(RM) training followed by LLM optimization via Proximal Policy Optimization~(PPO)~\citep{schulman2017proximalpolicyoptimizationalgorithms}. This paradigm has proven highly effective across academia and industry. Several subsequent work have been proposed methodical improvements to this foundation~\citep{rafailov2024directpreferenceoptimizationlanguage, azar2023generaltheoreticalparadigmunderstand, ethayarajh2024ktomodelalignmentprospect, shao2024deepseekmathpushinglimitsmathematical, hong2024orpomonolithicpreferenceoptimization, chen2024noise}. Direct Preference Optimization~(DPO)~\citep{rafailov2024directpreferenceoptimizationlanguage} reformulates reward functions to bypass Reward Modeling and PPO phases. Identity Preference Optimization~(IPO)~\citep{azar2023generaltheoreticalparadigmunderstand} introduces regularization addressing DPO's overfitting tendencies. Kolmogorov-Smirnov Optimization~(KTO)~\citep{ethayarajh2024ktomodelalignmentprospect} requires only binary annotations, reducing data collection costs. Group-wise Relative Policy Optimization~(GRPO)~\citep{shao2024deepseekmathpushinglimitsmathematical} employs intra-group relative rewards, eliminating value function estimation dependencies. Our framework enhances multi-instruction following capabilities of reward models, thereby improving RLHF training outcomes. Furthermore, our framework is directly applicable to these algorithms above, providing a more robust enhancement of multi-instruction following capabilities of models.

\paragraph{Multi-Instruction Following} Multi-instruction following capability is crucial for LLMs to complete complex tasks accurately~\citep{lou2024largelanguagemodelinstruction}. Previous approaches primarily fall into the following three categories: 1) Human annotation methods~\citep{DatabricksBlog2023DollyV2, köpf2023openassistantconversationsdemocratizing}, which enhance capabilities through deliberate dataset construction via human annotations. Dolly~\citep{DatabricksBlog2023DollyV2} construct a collection of human-authored prompt-response pairs for LLM fine-tuning. However, these approaches are resource-intensive. 2) LLM-based multi-instruction data augmentation~\citep{xu2023wizardlmempoweringlargelanguage, hui2024smallerlanguagemodelsbetter, sun2024coniferimprovingcomplexconstrained, huang2025muscimprovingcomplexinstruction, he2024complexsimpleenhancingmulticonstraint, zhang2024iopoempoweringllmscomplex}, which leverage LLMs to generate or respond to complex instructions. Conifer~\citep{sun2024coniferimprovingcomplexconstrained} uses GPT-4~\citep{achiam2023gpt} to create verified multi-level instructions. MuSC~\citep{huang2025muscimprovingcomplexinstruction} employs target LLMs for instruction decomposition and recombination. However, since these approach rely on the outputs generated by LLMs, there exists the problem of inaccuracies in those generated dataset. 3) Programmatic multi-instruction data augmentation, which generates datasets by programmatically constructing additional instructions on existing preference datasets. LIFT~\citep{yuan2024followinglengthconstraintsinstructions} enhances length instruction adherence through the implementation of specialized datasets. However, these may invert semantic preferences and degrade model performance. Our approach employs programmatically verifiable instructions for construction and validation, ensuring data correctness while implementing question-oriented preference data to mitigate semantic loss.
\section{Method}
\label{sec:3}

\subsection{Preliminary}

In Reinforcement Learning from Human Feedback (RLHF), the Bradley-Terry (BT) model~\citep{bradley1952rank} is commonly employed to model human preferences. Specifically, we prompt a model $\pi$ to generate two distinct responses $y_i$ and $y_j$ to an identical human query input $x$, and collect human preferences regarding this pair, forming a preference data tuple $(x, y_w, y_l)$, where $y_w \succ y_l$ indicates that the former is preferred over the latter. Subsequently, the BT model posits the existence of a true and accurate human preference reward function $r^*(x,y)$, thereby enabling the modeling of human preferences between these two responses as follow:

\begin{equation}
    \label{eqa:dataset_bt_0}
    p^*(y_w \succ y_l | x) = \frac{\exp(r^*(x, y_w))}{\exp(r^*(x, y_w)) + \exp(r^*(x, y_l))}.
\end{equation}

Due to the inherent complexity of human preferences, such a reward function cannot typically be directly obtained. Consequently, a parameterized reward model $r_{\phi}$ is employed and optimized through methods such as maximum likelihood estimation to approximate true human preferences, as expressed in the following equation:

\begin{equation}
    \label{eqa:dataset_bt}
    \begin{aligned}
        \mathcal{L}_{\textit{BT}} = & -\underset{(x, y_w, y_l) \sim \mathcal{D}_{\textit{BT}}}{\mathbb{E}}[\log p_\phi(y_w \succ y_l | x)].
    \end{aligned}
\end{equation}

where $\mathcal{D}_{BT} = \{(x, y_w, y_l)\}$~\footnote{For clarity of notation, we omit the superscript $^{(i)}$ in subsequent discussions where no ambiguity exists.} represents the collected preference dataset. When the acquired data is both substantial and high in quality, the BT model can perform effectively in preference modeling tasks. 
Besides, Response-Conditioned Bradley-Terry Model~(Rc-BT)~\citep{cai2025disentanglinglengthbiaspreference} focuses on preference comparisons between prompts, which may contain valuable latent signals. Specifically, this approach transforms prompt-conditioned modeling paradigm of traditional BT model into a response-conditioned modeling framework, thereby learning instruction-following capabilities more effectively. The formulation is expressed as follows:

\begin{equation}
    \label{eqa:rc_bt}
    \begin{aligned}
        p^*(x_w \succ x_l | y) = \frac{\exp(r^*(x_w, y))}{\exp(r^*(x_w, y)) + \exp(r^*(x_l, y))}.
    \end{aligned}    
\end{equation}

According to the aforementioned formulation, we can also optimize a parameterized reward model $r_\phi$ to approximate true human preferences through methods such as maximum likelihood estimation, as demonstrated in the following equation:

\begin{equation}
    \label{eqa:rc_bt_loss}
    \begin{aligned}
        \mathcal{L}_{\textit{Rc-BT}} = -\underset{(x_w, x_l, y) \sim \mathcal{D}_{\textit{Rc-BT}}}{\mathbb{E}}[\log p_\phi(x_w \succ x_l | y)].
    \end{aligned}    
\end{equation}

\subsection{Intra-Sample Level Preference Learning}
\label{sec:3.2}

As we mentioned in Section~\ref{sec:1}, the latent signals within prompts in preference data are also extremely valuable. To leverage these potentially valuable signals embedded in prompts, we apply response-conditioned modeling framework to multi-instruction following tasks. We first define functions for generating instructions, then select a subset of functions, provide two sets of arguments that constitute preference relation to these functions to generate intra-sample level preference data, and finally propose a training objective function to capture this preference information. The detailed algorithmic procedure is delineated in Algorithm~\ref{algo:intra} within Appendix~\ref{apx:algo_3.2}. Specifically, our implementation encompasses the following steps:

First, we define a function set $G_0 = \{g_1, g_2, ..., g_n\}$, where each function generates a instruction~$ins$ based on given arguments $\{(y_i, True/False)\}$, indicating whether $y_i$ does satisfy the generated instruction $ins$. Instructions generated by different functions are mutually distinct. Algorithm~\ref{algo:length_func} shows an example of functions defined above.
 
\begin{wrapfigure}[19]{r}{0.54\textwidth}
\vspace{-0.4cm}
  \begin{minipage}{0.54\textwidth}
\begin{algorithm}[H]
  \caption{Function example, Python-like}
  \label{algo:length_func}
\begin{verbatim}
# input: arguments
# output: length instruction string
def g_i(arguments):
    template = "Your response should \
        contain at least {} words."
    l_min, l_max = 0, inf
    for y, condition in arguments:
        if condition == True:
            l_max = min(l_max, len(y))
        else:
            l_min = max(l_min, len(y))
    if l_min > l_max:
        return "" # no solution
    else:
        length = randint(l_min, l_max)
        return template.format(length)
\end{verbatim}
\end{algorithm}
\end{minipage}
\end{wrapfigure}

To construct multi-instruction prompt, given original preference data $(x, y_w, y_l)$, we randomly select multiple functions $G_1 = \{g_{i_1}, \dots, g_{i_k}\}$ from $G_0$ and combine their outputs with original prompt $x$ to form a multi-instruction prompt $x'$.
To construct preference data for Eqn.~\ref{eqa:rc_bt}, we need to provide distinct sets of arguments to each function within $G_1$ to generate multi-instruction constraints and construct two multi-instruction prompts $x'_w$ and $x'_l$, that satisfy either of the following intra-sample level preference relationships:
1) The response $y$ conforms to all instructions in $x'_w$, , but does not conform to at least one instruction in $x'_l$.
2) The response $y$ does not conform to any instruction in $x'_l$, but conforms to at least one instruction in $x'_w$.
For other relationships, we cannot determine the preference between the two prompts, therefore we do not utilize them. For a given $y$, we randomly select one of the above two relationships to generate intra-sample level preference data $(x_w', x_l', y)$.

For the original preference data $(x, y_w, y_l)$, there exist two responses $y_w$ and $y_l$. To avoid introducing asymmetric bias, we generated one multi-instruction preference data for each of $y_w$ and $y_l$, denoted as $(x^1_w, x^1_l, y_w)$ and $(x^2_w, x^2_l, y_l)$. Then substituting $(x^1_w, x^1_l, y_w)$ and $(x^2_w, x^2_l, y_l)$ into Eqn.~\ref{eqa:rc_bt}, we derive the following intra-sample level preference modeling expression:

\begin{equation}
    \label{eqa:dataset_ours_bt}
    \begin{aligned}
        & p^*(x_w^1 \succ x_l^1 | y_w) = \frac{\exp{(r^*(x_w^1, y_w))}}{\exp(r^*(x_w^1, y_w))+\exp(r^*(x_l^1, y_w))}, \\
        & p^*(x_w^2 \succ x_l^2 | y_l) = \frac{\exp{(r^*(x_w^2, y_l))}}{\exp(r^*(x_w^2, y_l))+\exp(r^*(x_l^2, y_l))}.
    \end{aligned}
\end{equation}

Ultimately, our intra-sample level preference dataset is defined as $\mathcal{D}_{intra} = \{(x_w^1, x_l^1, y_w)\} \cup \{(x_w^2, x_l^2, y_l)\}$. Accordingly, we propose the following training objective function to capture preference differences at the intra-sample level from Eqn.~\ref{eqa:dataset_ours_bt}:

\begin{equation}
    \label{eqa:dataset_ours}
    \begin{aligned}
        \mathcal{L}_{\textit{intra}} = & -\underset{(x_w^1, x_l^1, y_w) \sim \mathcal{D}_{\textit{intra}}}{\mathbb{E}}[\log p_\phi(x_w^1 \succ x_l^1 | y_w)] \\
        & - \underset{(x_w^2, x_t^2, y_l) \sim \mathcal{D}_{\textit{intra}}}{\mathbb{E}}[\log p_\phi(x_w^2 \succ x_l^2 | y_l)].
    \end{aligned}    
\end{equation}

\subsection{Inter-Sample Level Preference Learning}
\label{sec:3.3}

Beyond preference distinctions within preference data, there also exists valuable preference differences between preference data. To capture these inter-sample level preference differences between preference data, we construct multi-instruction preference data pairs and compare them with original preference data pair $(x, y_w, y_l)$. We enumerate two different inter-sample level preference relationships and generate a distinct multi-instruction preference pair for each relationship, utilizing a subset of the function set $G_0$ defined in Section~\ref{sec:3.2}. These two multi-instruction preference pairs, each combined with the original data pair, constitute two distinct types of inter-sample level preference data. Finally, we propose a training objective function to capture the inter-sample level preference differences. The detailed algorithmic procedure is outlined in Algorithm~\ref{algo:inter} in Appendix~\ref{apx:algo_3.3}. Specifically, we implement the following procedure:

First, We randomly select functions $G_1 = \{g_{j_1}, \dots, g_{j_m}\}$ from the function set $G_0 = \{g_1, g_2, ..., g_n\}$, provide arguments $\{(y_w, True/False)$, $(y_l, True/False)\}$ for each $g_j$, and combine all returned instructions with $x$ to generate a multi-instruction prompt $x'$, forming $(x', y_w, y_l)$.

The inter-sample level preference relationships are shown as follows:
1) $y_w$ conforms to all instructions in $x'$, while $y_l$ does not conform to at least one instruction, resulting in $p^*(x', y_w, y_l) > p^*(x, y_w, y_l)$;
2) $y_w$ does not conform to any instruction in $x'$, while $y_l$ conforms to at least one instruction, resulting in $p^*(x', y_w, y_l) < p^*(x, y_w, y_l)$.
If both $y_w$ and $y_l$ partially conform to the instructions in $x'$, the preference relationship cannot be determined.

Based on these relationships and selected functions, we generate two multi-instruction preference data with preference relationships: $(x_w, y_w, y_l)$ and $(x_l, y_w, y_l)$, where $p^*(x_w, y_w, y_l) > p^*(x, y_w, y_l)$ and $p^*(x_l, y_w, y_l) < p^*(x, y_w, y_l)$. To model the preference described above using Eqn.~\ref{eqa:dataset_bt_0}, we utilize logit function $\gamma(p) = \log(\frac{p}{1-p})$ to transform $p^*$ into logits, then extend the BT model to inter-sample level preference modeling, resulting in the following expression:
\begin{equation}
    \label{eqa:dataset_gap_bt}
    \begin{aligned}
        & p^*((x_w, y_w, y_l) \succ (x, y_w, y_l)) = \frac{\exp( \gamma(p^*(y_w \succ y_l | x_w)))}{\exp(\gamma(p^*(y_w \succ y_l | x_w))) + \exp(\gamma(p^*(y_w \succ y_l | x)))}, \\
        & p^*((x, y_w, y_l) \succ (x_l, y_w, y_l)) = \frac{\exp(\gamma(p^*(y_w \succ y_l | x)))}{\exp(\gamma(p^*(y_w \succ y_l | x))) + \exp(\gamma(p^*(y_w \succ y_l | x_l)))}.
    \end{aligned}
\end{equation}

Ultimately, our inter-sample level preference dataset is defined as $\mathcal{D}_{inter} = \{(x_w, x, y_w, y_l)\} \cup \{(x, x_l, y_w, y_l)\}$. Accordingly, we propose the following training objective function to capture preference differences at the inter-sample level derived from Eqn.~\ref{eqa:dataset_gap_bt}:

\begin{equation}
    \label{eqa:dataset_gap}
    \begin{aligned}
        \mathcal{L}_{\textit{inter}} = & -\underset{(x_w, x, y_w, y_l) \sim \mathcal{D}_{\textit{inter}}}{\mathbb{E}}[\log p_\phi((x_w, y_w, y_l) \succ (x, y_w, y_l))] \\
        & - \underset{(x, x_l, y_w, y_l) \sim \mathcal{D}_{\textit{inter}}}{\mathbb{E}}[\log p_\phi((x, y_w, y_l) \succ (x_l, y_w, y_l))].
    \end{aligned}
\end{equation}

Finally, in \ourm{}, we simultaneously employ Eqn.~\ref{eqa:dataset_bt},~\ref{eqa:dataset_ours}, and~\ref{eqa:dataset_gap} to model data preferences. Consequently, our \ourm{} dataset is defined as $\mathcal{D}_\textit{MAPL} = \{(x, y_w, y_l)\} \cup \{(x_w^1, x_l^1, y_w)\} \cup \{(x_w^2, x_l^2, y_l)\} \cup \{(x_w, x, y_w, y_l)\} \cup \{(x, x_l, y_w, y_l)\}$, with the preference modeling expression formulated as follows:

\begin{equation}
    \label{eqa:dataset_ours+gap}
    \begin{aligned}
        \mathcal{L}_\textit{MAPL} = \mathcal{L}_\textit{BT} + \mathcal{L}_\textit{intra} + \mathcal{L}_\textit{inter}.
    \end{aligned}    
\end{equation}

\subsection{Multi-Level Aware Reward Modeling and Direct Preference Optimization}
\label{sec:3.4}

In this subsection, we derive the application of our framework to reward modeling and DPO. Complete derivation formulas can be found in the Appendix~\ref{apx:proof_all}. 
For Eqn.~\ref{eqa:dataset_bt} and~\ref{eqa:dataset_ours}, through maximum likelihood estimation, employing a parameterized reward model $r_\phi$, we obtain the following negative log-likelihood loss for intra-sample level preference modeling:

\begin{equation}
    \label{eqn:rm_loss_BT}
    \begin{aligned}
        \mathcal{L}_{r_\phi}(\mathcal{D}_{\textit{BT}}) = & -\underset{(x, y_w, y_l) \sim \mathcal{D}_{\textit{MAPL}}}{\mathbb{E}}[\log\sigma (r_\phi(x, y_w) - r_\phi(x, y_l))],
    \end{aligned}
\end{equation}

\begin{equation}
    \label{eqn:rm_loss_ours}
    \begin{aligned}
        \mathcal{L}_{r_\phi}(\mathcal{D}_{\textit{intra}}) = & - \underset{(x_w^1, x_l^1, y_w) \sim \mathcal{D}_{\textit{MAPL}}}{\mathbb{E}}[\log\sigma (r_\phi(x_w^1, y_w) - r_\phi(x_l^1, y_w))] \\
        & - \underset{(x_w^2, x_l^2, y_l) \sim \mathcal{D}_{\textit{MAPL}}}{\mathbb{E}}[\log\sigma (r_\phi(x_w^2, y_l) - r_\phi(x_l^2, y_l))].
    \end{aligned}
\end{equation}

For Eqn.~\ref{eqa:dataset_gap}, through maximum likelihood estimation, employing a parameterized model $r_\phi$, we obtain the following negative log-likelihood loss for inter-sample level preference modeling:

\begin{equation}
    \label{eqn:rm_loss_gap}
    \begin{aligned}
        \mathcal{L}_{r_\phi}(\mathcal{D}_{\textit{inter}}) = & - \underset{(x_w, x, y_w, y_l) \sim \mathcal{D}_{\textit{MAPL}}}{\mathbb{E}}[\log\sigma ((r_\phi(x_w, y_w) - r_\phi(x_w, y_l)) - (r_\phi(x, y_w) - r_\phi(x, y_l)))] \\
        & - \underset{(x, x_l, y_w, y_l) \sim \mathcal{D}_{\textit{MAPL}}}{\mathbb{E}}[\log\sigma ((r_\phi(x, y_w) - r_\phi(x, y_l)) - (r_\phi(x_l, y_w) - r_\phi(x_l, y_l)))].
    \end{aligned}
\end{equation}

Ultimately, the loss function for our \ourm{} framework applied to reward modeling is formulated as:

\begin{equation}
    \label{eqa:rm_loss_ours+gap}
    \begin{aligned}
        \mathcal{L}_{r_\phi}(\mathcal{D}_{\textit{MAPL}}) = \mathcal{L}_{r_\phi}(\mathcal{D}_{\textit{BT}}) + \mathcal{L}_{r_\phi}(\mathcal{D}_{\textit{intra}}) + \mathcal{L}_{r_\phi}(\mathcal{D}_{\textit{inter}}).
    \end{aligned}    
\end{equation}

Meanwhile, direct preference modeling is achieved through derivations of RM and RL objective. According to the derivation of Appendix A in DPO~\cite{rafailov2024directpreferenceoptimizationlanguage} and Appendix B in Rc-BT~\cite{cai2025disentanglinglengthbiaspreference}, by expressing the reward function as a function of the optimal policy and substituting it into Eqn.~\ref{eqa:dataset_bt} and~\ref{eqa:dataset_ours} respectively, we derive the expressions for original DPO and Intra-sample Level DPO as follows:

\begin{equation}
    \label{eqa:dpo_loss_BT+intra}
    \begin{aligned}
        & \mathcal{L}_{\textit{DPO}}^{\textit{BT}}(\pi_\theta; \pi_\text{ref}) = -\underset{(x, y_w, y_l) \sim \mathcal{D}_{\textit{MAPL}}}{\mathbb{E}} [\log\sigma(\beta \log\frac{\pi_\theta(x, y_w)}{\pi_\text{ref}(x, y_w)} - \beta \log\frac{\pi_\theta(x, y_l)}{\pi_\text{ref}(x, y_l)})], \\ & \mathcal{L}_{\textit{DPO}}^{\textit{intra}}(\pi_\theta; \pi_\text{ref}) = \\ & -\underset{(x_w^1, x_l^1, y_w) \sim \mathcal{D}_{\textit{MAPL}}}{\mathbb{E}} [\log\sigma(\beta \log\frac{\pi_\theta(x_w^1, y_w)}{\pi_\text{ref}(x_w^1, y_w)} - \beta \log\frac{\pi_\theta(x_l^1, y_w)}{\pi_\text{ref}(x_l^1, y_w)})] \\ & -\underset{((x_w^2, x_l^2, y_l) \sim \mathcal{D}_{\textit{MAPL}}}{\mathbb{E}} [\log\sigma(\beta \log\frac{\pi_\theta(x_w^2, y_l)}{\pi_\text{ref}(x_w^2, y_l)} - \beta \log\frac{\pi_\theta(x_l^2, y_l)}{\pi_\text{ref}(x_l^2, y_l)})].
    \end{aligned}
\end{equation}

Similarly, by incorporating the preference probabilities $p^*(y_w \succ y_l | x)$ and $p^*(y_w \succ y_l | x_w)$ (or $p^*(y_w \succ y_l | x_l)$) from Eqn.~\ref{eqa:dataset_gap}, and substituting the reward function expressed as a function of the optimal policy, we obtain the expression for Inter-sample Level DPO as follows:

\begin{equation}
    \label{eqa:dpo_loss_BT+inter}
    \begin{alignedat}{2}
        & \mathcal{L}_{\textit{DPO}}^{\textit{inter}}(\pi_\theta; \pi_\text{ref}) = \\
        & -\underset{(x_w, x, y_w, y_l) \sim \mathcal{D}_{\textit{MAPL}}}{\mathbb{E}}[\log\sigma( \,
        && (\beta \log\frac{\pi_\theta(y_w \mid x_w)}{\pi_\text{ref}(y_w \mid x_w)} - \beta \log\frac{\pi_\theta(y_l \mid x_w)}{\pi_\text{ref}(y_l \mid x_w)}) \\
        &&& - (\beta \log\frac{\pi_\theta(y_w \mid x)}{\pi_\text{ref}(y_w \mid x)} - \beta \log\frac{\pi_\theta(y_l \mid x)}{\pi_\text{ref}(y_l \mid x)}))] \\
        & -\underset{(x, x_l, y_w, y_l) \sim \mathcal{D}_{\textit{MAPL}}}{\mathbb{E}} [\log\sigma(\,
        && (\beta \log\frac{\pi_\theta(y_w \mid x)}{\pi_\text{ref}(y_w \mid x)} - \beta \log\frac{\pi_\theta(y_l \mid x)}{\pi_\text{ref}(y_l \mid x)}) \\
        &&& - (\beta \log\frac{\pi_\theta(y_w \mid x_l)}{\pi_\text{ref}(y_w \mid x_l)} - \beta \log\frac{\pi_\theta(y_l \mid x_l)}{\pi_\text{ref}(y_l \mid x_l)}))].
    \end{alignedat}
\end{equation}

Ultimately, the loss function for our \ourm{} framework applied to DPO is formulated as:

\begin{equation}
    \label{eqa:dpo_loss_ours+gap}
    \begin{aligned}
        & \mathcal{L}_{\textit{DPO}}^{\textit{MAPL}}(\pi_\theta; \pi_\text{ref}) = \mathcal{L}_{\textit{DPO}}^{\textit{BT}}(\pi_\theta; \pi_\text{ref}) + \mathcal{L}_{\textit{DPO}}^{\textit{intra}}(\pi_\theta; \pi_\text{ref}) + \mathcal{L}_{\textit{DPO}}^{\textit{inter}}(\pi_\theta; \pi_\text{ref}).
    \end{aligned}
\end{equation}
\section{Experiments}
\label{sec:4}

In this section, we conduct comprehensive experimental evaluations to demonstrate the efficacy of our proposed \ourm{} framework. Additionally, we perform ablation studies to validate the effectiveness of key design components in our approach.

\subsection{Experimental Settings}
\label{sec:4.1}

\paragraph{Dataset and Models} For our dataset, 
we extract the first turn of English conversations from the OpenAssistant dataset \citep{köpf2023openassistantconversationsdemocratizing} and based on their human-annotated rankings, we designate rank $0$ as "chosen" ($y_w$) and rank $1$ as "rejected" ($y_l$), thereby generating the origin preference dataset $\mathcal{D} = \{(x, y_w, y_l)\}$. 
We extract a subset from $\mathcal{D}$ to create dataset $\mathcal{D}_\textit{BT}$, with the remaining portion serving as the semantic evaluation dataset $\mathcal{D}_\textit{eval}$. Subsequently, we construct a multi-instruction evaluation dataset $\mathcal{D}_\textit{eval}^\textit{if}$ based on $\mathcal{D}_\textit{eval}$ to assess capacities for multi-instruction following. The detailed construction process is described in Appendix~\ref{apx:reward_model_eval}. 
As mentioned in Section~\ref{sec:3}, we generate dataset $\mathcal{D}_{\textit{MAPL}}$ based on $\mathcal{D}_{\textit{BT}}$.
RM and DPO models are both trained on $\mathcal{D}_{\textit{MAPL}}$, designated as MAPL-RM and MAPL-DPO respectively. For our model architecture, we employ three models as foundation models: Qwen2-1.5B-Instruct, Qwen2-7B-Instruct~\citep{yang2024qwen2technicalreport}, and Llama-3.1-8B-Instruct~\citep{grattafiori2024llama3herdmodels}.

\paragraph{Training Details} For our MAPL-RM training, we set the learning rate to $9.65 \times 10^{-6}$, followed by a cosine learning rate schedule with initial warm-up for $10$ steps, and a batch size of $256$. Each experiment is trained for $5$ epochs. For MAPL-DPO training, except for the learning rate of $9.65 \times 10^{-7}$ and batch size of $8$, other settings remain the same as the RM training. All experiments are implemented using DeepSpeed~\citep{aminabadi2022deepspeedinferenceenablingefficient}, DeepSpeed-Chat~\citep{yao2023deepspeedchateasyfastaffordable} and Huggingface Transformers~\citep{wolf-etal-2020-transformers}, and conducted on a machine equipped with $8$ NVIDIA A$100$ GPUs, each with $80$GB of memory.

\paragraph{Compared Methods and Evaluations} We use models trained on $\mathcal{D}_\textit{BT}$ as baselines, denoted as Vanilla~RM and Vanilla~DPO. To validate the superiority of our framework, we conduct comparative analyses against M-LIFT\footnote{Multi-instruction LIFT (M-LIFT) applies LIFT~\cite{yuan2024followinglengthconstraintsinstructions} to multi-instruction following tasks. Our implemented M-LIFT method is detailed in Appendix~\ref{apx:M-LIFT_implement}.} and MuSC\footnote{MuSC necessitates specific datasets; therefore, we utilize the dataset provided by MuSC for training.}~\citep{huang2025muscimprovingcomplexinstruction}. For Reward Model evaluation, we employ two principal metrics to assess performance across different methods: {\em Semantic Quality Eval Acc} (SQ) and {\em Instruction Following Eval Acc} (IF), which represent accuracy rates on $\mathcal{D}_\textit{eval}$ and $\mathcal{D}_\textit{eval}^\textit{if}$ respectively. The former metric evaluates semantic quality, while the latter measures multi-instruction following capacity. IFEval~\citep{zhou2023instructionfollowingevaluationlargelanguage} is a reproducible benchmark designed to quantitatively assess how well LLMs follow multi-instructions by focusing on $25$ types of verifiable instructions across approximately $500$ prompts. For DPO evaluation, we utilize the included {\em Prompt-level Strict Acc} (Prompt) and {\em Instruction-level Strict Acc} (Inst) metrics to assess performance across methods. MT-Bench~\citep{zheng2023judging} is a challenging multi-turn benchmark that assesses LLMs’ performance across different tasks via strong LLM evaluators acting as scalable judges aligned with human preferences. We utilize the {\em Average Score} of the first turn and the second turn to assess the semantic quality of models.

\subsection{Main Results}

\begin{table*}[!ht]
    \centering
    \caption{Experimental results across three models implementing different methods, evaluated on {\em Semantic Quality Evaluation Accuracy} (SQ) and {\em Instruction Following Evaluation Accuracy} (IF).}
    \begin{tabular}{ccccccc}
        \toprule
        \multirow{2}{*}{Methods} &\multicolumn{2}{c}{\textbf{Qwen2-1.5B-Instruct}}&\multicolumn{2}{c}{\textbf{Qwen2-7B-Instruct}}&\multicolumn{2}{c}{\textbf{Llama-3.1-8B-Instruct}} \\
        \cmidrule(lr){2-3} \cmidrule(lr){4-5} \cmidrule(lr){6-7} & IF & SQ & IF & SQ & IF & SQ \\
        \midrule
        Vanilla RM&64.17&60.56&68.75&61.94&63.75&62.22 \\
        M-LIFT&77.22&60.00&81.67&60.56&82.92&61.67 \\
        MuSC&69.03&55.28&72.50&61.39&75.28&60.28 \\
        Ours&{\bf 91.81}&{\bf 61.94}&{\bf 94.31}&{\bf 63.89}&{\bf 94.31}&{\bf 63.33} \\ 
        \bottomrule
    
    \end{tabular}
    \label{table:RM_ifeval_result}
\end{table*}

\begin{table*}[!ht]
    \centering
    \caption{Experimental results comparing different methods across three models on IFEval metrics. Prompt is {\em Prompt-level Strict Acc} and Inst is {\em Instruction-level Strict Acc}.}
    \begin{tabular}{ccccccc}
        \toprule
        \multirow{2}{*}{Methods} &\multicolumn{2}{c}{\textbf{Qwen2-1.5B-Instruct}}&\multicolumn{2}{c}{\textbf{Qwen2-7B-Instruct}}&\multicolumn{2}{c}{\textbf{Llama-3.1-8B-Instruct}} \\
        \cmidrule(lr){2-3} \cmidrule(lr){4-5} \cmidrule(lr){6-7} & Prompt & Inst & Prompt & Inst & Prompt & Inst \\
        \midrule
        Vanilla DPO&26.30&36.78&49.26&59.13&67.04&75.00 \\ 
        M-LIFT&35.00&46.15&56.30&64.54&70.40&78.52 \\
        MuSC&27.39&38.83&48.53&58.90&64.88&74.46 \\
        Ours&{\bf 42.65}&{\bf 54.77}&{\bf 57.54}&{\bf 66.11}&{\bf 71.51}&{\bf 79.24} \\ 
        \bottomrule
    
    \end{tabular}
    \label{table:DPO_ifeval_result}
\end{table*}

Table~\ref{table:RM_ifeval_result} presents our experimental results across two metrics (Semantic Quality Evaluation Accuracy (SQ) and Instruction Following Evaluation Accuracy (IF)). M-LIFT, MuSC, and our \ourm{} framework all demonstrate varying degrees of improvement in IF metric compared to Vanilla RM, with our framework achieving the most substantial enhancement. Specifically, using Qwen2-7B-Instruct as the base model, our framework surpasses M-LIFT by $12.64\%$ and MuSC by $21.81\%$ in IF metrics, with comparable improvements observed for the other two models (Qwen2-1.5B-Instruct and Llama3.1-8B-Instruct). Concurrently, in terms of semantic quality of model outputs, e.g., SQ metric, also with Qwen2-7B-Instruct as the base model, M-LIFT exhibits a $1.38\%$ degradation and MuSC exhibits a $0.55\%$ degradation compared to Vanilla RM, whereas our framework maintains semantic quality without noticeable degradation. Similar patterns of performance are observed across Qwen2-1.5B-Instruct and Llama3.1-8B-Instruct models.

Table~\ref{table:DPO_ifeval_result} presents our experimental results on two metrics from the instruction-following benchmark IFEval (Prompt-level Strict Acc and Instruction-level Strict Acc). M-LIFT, MuSC, and our \ourm{} framework all demonstrate varying degrees of improvement in multi-instruction following capabilities compared to Vanilla DPO, with our approach achieving the most substantial enhancement. Using Qwen2-7B-Instruct as the base model, our framework surpasses M-LIFT by $1.24\%$ and MuSC by $9.01\%$ on Prompt-level Strict Acc, while exceeding M-LIFT by $1.57\%$ and MuSC by $7.21\%$ on Instruction-level Strict Acc, with comparable performance patterns observed for the other two models.

Our framework avoid using LLMs for instruction generation, instead employing programmatic approaches for compliance assessment and instruction generation, thereby ensuring accuracy of~generated outcomes and consequently achieving superior multi-instruction following capabilities. Furthermore, our framework not only addresses preference within preference data but also models between preference pairs, thereby achieving superior enhancement in multi-instruction following capabilities compared to alternative methods. Experimental results shown above confirm our conclusions.

\begin{wraptable}[10]{r}{0.5\textwidth}
    \vspace{-0.3cm}
    \caption{Experimental results comparing different methods across models on MT-Bench metrics.}
    \centering
    \begin{tabular}{ccc}
        \toprule
        \multirow{2}{*}{Methods} &{\textbf{Qwen2-7B}}&{\textbf{Llama-3.1-8B}} \\
        & {\textbf{Instruct}} & {\textbf{Instruct}} \\
        \midrule
        Vanilla DPO&{\bf 7.26}&7.73 \\ 
        M-LIFT&6.99&7.45 \\
        Ours&7.22&{\bf 7.75} \\ 
        \bottomrule
    \end{tabular}
    \label{table:DPO_mtbench_result}
\end{wraptable}

Besides, as mentioned in Section~\ref{sec:1}, certain methods of instruction data augmentation that generate multi-instruction prompt for original dataset, such as M-LIFT, can damage the semantic quality of model outputs. Our framework designs a new intra-sample level preference data that avoids reversing semantic preference order in the original dataset in Section~\ref{sec:3.2}. Also, the inter-sample level preference data we introduced in Section~\ref{sec:3.3} does not reverse the semantic preference order of the original dataset, thereby preventing degradation in the semantic quality of model outputs. Experimental results shown in Table~\ref{table:DPO_mtbench_result} also confirm our conclusions. Specifically, compared to Vanilla DPO, M-LIFT resulted in a $0.27$ drop in MT-Bench Average Score when training with Qwen2-7B-Instruct, with similar decreases observed in other models. In contrast, models trained using our framework showed no score reduction compared to Vanilla DPO.

\begin{figure*}[!ht]
    \includegraphics[width=\textwidth]{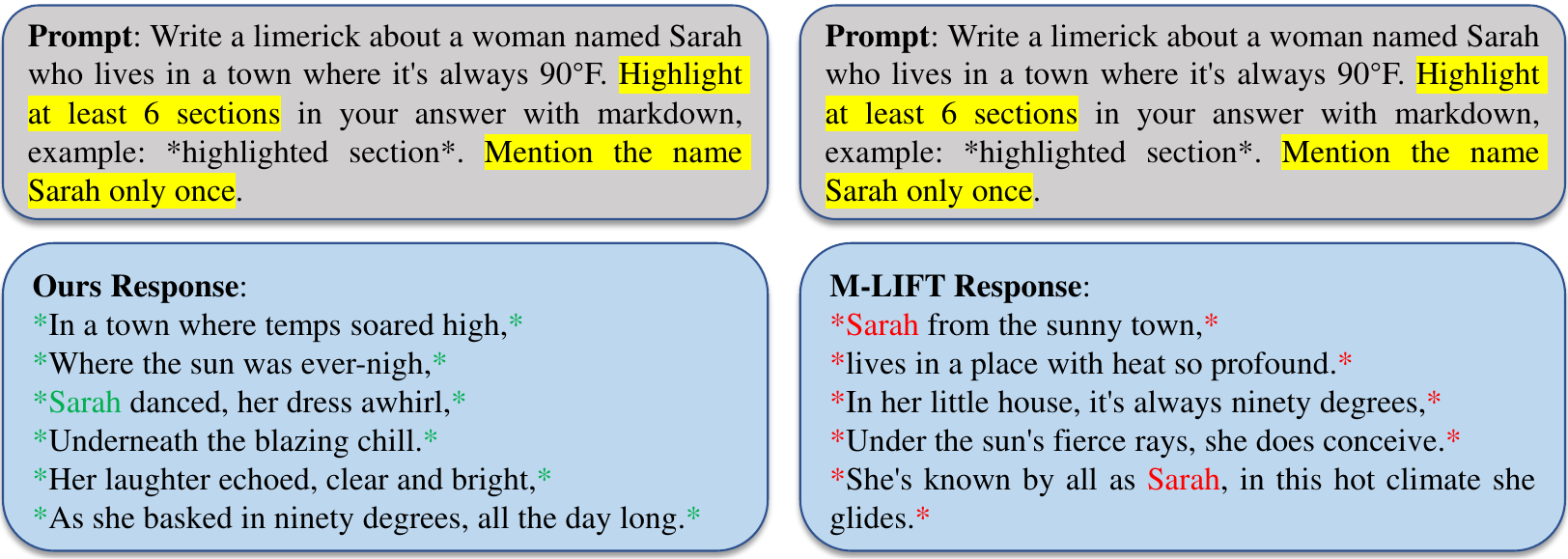}
    \caption{Case study on models based on Qwen2-7B-Instruct and trained through our framework and M-LIFT.}
    \label{fig:case_study}
\end{figure*}

\paragraph{Case Study on DPO.} Figure~\ref{fig:case_study} presents a case study demonstrating the superiority of MAPL-DPO training compared to other comparative methods. We conduct inference using models based on Qwen2-7B-Instruct and trained through our framework and M-LIFT. The prompt imposes three constraints: "highlight 6 sections with **", "mention the name Sarah" and "mention the name Sarah only once". Ours response successfully satisfies all constraints. Conversely, M-LIFT's response fails to meet two constraints ("highlight at least 6 sections" and "mention the name Sarah only once") with error locations marked in red. This demonstrates that our framework, which leverages both intra-sample level and inter-sample level preference information, exhibits superior multi-instruction following capabilities compared to M-LIFT.

\subsection{Ablation Studies}

\begin{wraptable}[13]{r}{0.52\textwidth}
    \centering
    \caption{Ablation study results of model based on Qwen2-7B-Instruct and trained with different methods, evaluated on IFEval.}
    \renewcommand{\arraystretch}{1.2}
    \begin{tabular}{ccc}
        \toprule
        \multirow{2}{*}{Methods} &\multicolumn{2}{c}{\textbf{Qwen2-7B-Instruct}} \\
        \cmidrule(lr){2-3} & Prompt & Inst \\
        \midrule
        Vanilla DPO&49.26&59.13 \\ 
        only Intra-sample Level&56.25&65.99 \\
        only Inter-sample Level&54.04&62.53 \\
        \ourm{} (Ours)&{\bf 57.54}&{\bf 66.11} \\ 
        \bottomrule
    
    \end{tabular}
    \label{tab:ablation}
\end{wraptable}

We conduct an ablation study to examine the impact of intra-sample level preference learning and inter-sample level preference learning proposed in Section~\ref{sec:3.2} and~\ref{sec:3.3}, using Qwen2-7B-Instruct as the base model and training with DPO. As illustrated in Table~\ref{tab:ablation}, models trained with either part of MAPL consistently demonstrate superior performance compared to Vanilla DPO. However, when compared to our comprehensive approach (incorporating both parts), eliminating inter-sample level preference learning results in modest performance decrements on IFEval metrics, whereas removing intra-sample level preference learning leads to more substantial performance degradation. We attribute this performance decrement to the complementary nature of the two parts of \ourm{} framework: inter-sample level preference learning and intra-sample level preference learning focus on preference within and between preference pairs respectively, thereby generating mutually reinforcing effects that yield superior results compared to any singular part.
\section{Conclusion and Future Work}
\label{sec:5}

This paper primarily investigates methods for enhancing multi-instruction following capabilities in LLMs. We propose~\ourmlong{}, a novel preference learning framework which incorporates intra-sample level preference learning within preference data pairs and inter-sample level preference learning between preference data pairs, avoiding the use of advanced LLMs or reversing original semantic preferences, thereby enhancing multi-instruction capabilities. Experiments on RM and DPO with multiple benchmarks demonstrate the efficacy of our framework. 

For future work, there are many interesting directions. First, we will investigate how to reduce computational cost of training within the MAPL framework. Second, we can extend the function set to include other non-programmatically verifiable instructions to enhance model's ability to follow more complex instructions. Third, how to apply multi-level preference information in other preference alignment tasks is also a promising research direction.

\section*{Limitations}
\label{sec:limitation}

Despite these strong gains, our approach has several limitations. First, it expands a single original preference data into multiple multi-instruction preference data, thereby increasing computational requirements. Besides, generalization capabilities across non-programmatically verifiable instructions remain unexplored, representing a primary direction for our subsequent research endeavors.

\bibliographystyle{plain}
\bibliography{main}


\newpage
\appendix

\section{Additional Details of MAPL Framework}
\label{apx:algo}

\subsection{Intra-Sample Level Preference Data Generation}
\label{apx:algo_3.2}

In this subsection, we present the algorithmic procedure for intra-sample level preference learning referenced in Section \ref{sec:3.2} as follows:

\begin{algorithm}[H]
\label{algo:intra}
\SetAlgoLined
\KwIn{Function set $G_0 = \{g_1, g_2, \ldots, g_n\}$, Original preference data $\mathcal{D}_{BT} = \{(x, y_w, y_l)\}$}
\KwOut{Intra-sample level preference data $\mathcal{D}_{intra} = \{(x^1_w, x^1_l, y_w)\} \cup \{(x^2_w, x^2_l, y_l)\}$}

\SetKwProg{Proc}{Procedure}{:}{}
\Proc{Generate($x, y, pattern, target$)}{
    Select subset of functions $\{g_{i_1}, g_{i_2}, \ldots, g_{i_k}\}$ from $G_0$\;
    $\mathcal{I} \leftarrow \emptyset$\;
    
    \uIf{$pattern = \text{"all"}$}{
        \ForEach{$g_{i_k}$}{
            $condition \leftarrow (y, target)$\;
            $instruction \leftarrow g_{i_k}(condition)$\;
            $\mathcal{I} \leftarrow \mathcal{I} \cup \{instruction\}$\;
        }
    }
    \uElseIf{$pattern = \text{"at\_least\_one"}$}{
        Randomly select at least one function $\{g_{j_k}\}$ from selected functions\;
        \ForEach{$g_{i_k}$}{
            \eIf{$g_{i_k}$ in $\{g_{j_k}\}$}{
                $condition \leftarrow (y, target)$\;
            }{
                $condition \leftarrow (y, \neg\,target)$\;
            }
            $instruction \leftarrow g_{i_k}(condition)$\;
            $\mathcal{I} \leftarrow \mathcal{I} \cup \{instruction\}$\;
        }
    }
    
    $x' \leftarrow$ Combine $x$ with $\mathcal{I}$\;
    \KwRet $x'$\;
}

\tcp{Main procedure}
$\mathcal{D}_{intra} \leftarrow \emptyset$\;
\ForEach{$(x, y_w, y_l)$ in $\mathcal{D}_{BT}$}{
    $x_w^1 \leftarrow$ Generate($x, y_w, \text{"all"}, True$)\;
    $x_l^1 \leftarrow$ Generate($x, y_w, \text{"at\_least\_one"}, False$)\;
    $x_w^2 \leftarrow$ Generate($x, y_l, \text{"at\_least\_one"}, True$)\;
    $x_l^2 \leftarrow$ Generate($x, y_l, \text{"all"}, False$)\;
    $\mathcal{D}_{intra} \leftarrow \mathcal{D}_{intra} \cup \{(x_w^1, x_l^1, y_w)\} \cup \{(x_w^2, x_l^2, y_l)\}$
}

\caption{Intra-Sample Level Preference Data Generation}
\end{algorithm}

\subsection{Inter-Sample Level Preference Data Generation}
\label{apx:algo_3.3}

In this subsection, we present the algorithmic procedure for inter-sample level preference learning referenced in Section \ref{sec:3.3} as follows:

\begin{algorithm}[H]
\label{algo:inter}
\SetAlgoLined
\KwIn{Function set $G_0 = \{g_1, g_2, \ldots, g_n\}$, Original preference data $\mathcal{D}_{BT} = (x, y_w, y_l)$}
\KwOut{Inter-sample level preference dataset $\mathcal{D}_{inter} = \{(x_w, x, y_w, y_l)\} \cup \{(x, x_l, y_w, y_l)\}$}

\SetKwProg{Proc}{Procedure}{:}{}
\Proc{Generate($x, y_w, y_l, target$)}{
    Select subset of functions $\{g_{i_1}, g_{i_2}, \ldots, g_{i_k}\}$ from $G_0$\;
    $\mathcal{I} \leftarrow \emptyset$\;
    
        Randomly select at least one function $\{g_{j_k}\}$ from selected functions\;
        \ForEach{$g_{i_k}$}{
            $condition_w \leftarrow (y_w, target)$\;
            
            \eIf{$g_{i_k}$ in $\{g_{j_k}\}$}{
                $condition_l \leftarrow (y_l, \neg\,target)$\;
            }
            {
                $condition_l \leftarrow (y_l, target)$\;
            }
            $instruction \leftarrow g_i(condition_w \cup condition_l)$\;
            
            $\mathcal{I} \leftarrow \mathcal{I} \cup \{instruction\}$\;
        }
    
    $x' \leftarrow$ Combine $x$ with $\mathcal{I}$\;
    \KwRet $x'$\;
}
\tcp{Main procedure}

$\mathcal{D}_{inter} \leftarrow \emptyset$\;
\ForEach{$(x, y_w, y_l)$ in $\mathcal{D}_{BT}$}{
    $x_w \leftarrow$ Generate($x, y_w, y_l, True$)\;
    $x_l \leftarrow$ Generate($x, y_w, y_l, False$)\;
    $\mathcal{D}_{inter} \leftarrow \mathcal{D}_{inter} \cup \{(x_w, x, y_w, y_l)\} \cup \{(x, x_l, y_w, y_l)\}$
}

\caption{Inter-Sample Level Preference Data Generation}
\end{algorithm}

\section{Derivation of Applying \ourm{} Framework to RM and DPO}
\label{apx:proof_all}

\subsection{Derivation of Inter-Sample Level RM Objective}

This subsection derives the inter-sample level RM objective. For notational convenience, we unify the two formulations from Eqn.~\ref{eqa:dataset_gap} in Section~\ref{sec:3.3} into the following representation:

\begin{equation}
    \label{epa:appx_b_inter_level_RM_bt}
    p^*((x_1, y_w, y_l) \succ (x_2, y_w, y_l)) \\ = \frac{\exp( \gamma(p^*(y_w \succ y_l | x_1)))}{\exp(\gamma(p^*(y_w \succ y_l | x_1))) + \exp(\gamma(p^*(y_w \succ y_l | x_2)))}. 
\end{equation}

Simultaneously, based on the BT model (Eqn.~\ref{eqa:dataset_bt_0}), we can derive:

\begin{equation}
    \begin{aligned}
    p^*(y_w\succ y_l|x) &= \frac{\exp{(r^*(x, y_w))}}{\exp(r^*(x, y_w))+\exp(r^*(x, y_l))} \\ &= \sigma(r^*(x, y_w) - r^*(x, y_l)).
    \label{eqa:appx_b_BT_RM_bt}
    \end{aligned}
\end{equation}

Substituting Eqn.~\ref{eqa:appx_b_BT_RM_bt} into Eqn.~\ref{epa:appx_b_inter_level_RM_bt}, we can derive:
\begin{equation}
    \label{eqa:appx_b_inter_level_RM_bt_derive}
    \begin{aligned}
        & p^*((x_1, y_w, y_l) \succ (x_2, y_w, y_l)) \\ &= \frac{\exp( \gamma(p^*(y_w \succ y_l | x_1)))}{\exp(\gamma(p^*(y_w \succ y_l | x_1))) + \exp(\gamma(p^*(y_w \succ y_l | x_2)))}. \\ &= \frac{\exp( \gamma(\sigma(r^*(x_1, y_w) - r^*(x_1, y_l))))}{\exp(\gamma(\sigma(r^*(x_1, y_w) - r^*(x_1, y_l)))) + \exp(\gamma(\sigma(r^*(x_2, y_w) - r^*(x_2, y_l))))} \\ &= \frac{\exp( r^*(x_1, y_w) - r^*(x_1, y_l))}{\exp(r^*(x_1, y_w) - r^*(x_1, y_l)) + \exp(r^*(x_2, y_w) - r^*(x_2, y_l))} \\ &= \frac{1}{1 + \exp(-[(r^*(x_1, y_w) - r^*(x_1, y_l)) - (r^*(x_2, y_w) - r^*(x_2, y_w))])} \\ &= \sigma((r^*(x_1, y_w) - r^*(x_1, y_l)) - (r^*(x_2, y_w) - r^*(x_2, y_w))).
    \end{aligned}
\end{equation}

Finally, we derive the loss function for inter-sample level RM for a parameterized policy $r_\phi$ using the inter-sample level preference dataset $\mathcal{D}_\textit{inter} = \{(x_w, x, y_w, y_l)\} \cup \{(x, x_l, y_w, y_l)\}$ by applying maximum likelihood estimation:

\begin{equation}
    \label{eqn:appx_B_rm_loss_gap}
    \begin{aligned}
        \mathcal{L}_{r_\phi}(\mathcal{D}_{\textit{inter}}) = & - \underset{(x_w, x, y_w, y_l) \sim \mathcal{D}_{\textit{inter}}}{\mathbb{E}}[\log\sigma ((r_\phi(x_w, y_w) - r_\phi(x_w, y_l)) - (r_\phi(x, y_w) - r_\phi(x, y_l)))] \\
        & - \underset{(x, x_l, y_w, y_l) \sim \mathcal{D}_{\textit{inter}}}{\mathbb{E}}[\log\sigma ((r_\phi(x, y_w) - r_\phi(x, y_l)) - (r_\phi(x_l, y_w) - r_\phi(x_l, y_l)))].
    \end{aligned}
\end{equation}

\subsection{Derivation of Inter-Sample Level DPO Objective}

According to Eqn.~5 in the DPO paper, we derive the expression of the reward function in terms of policy as follows:

\begin{equation}
    \begin{aligned}
    \label{eqa:appx_B_DPO_rstar}
        r^*(x, y) = \beta \log \frac{\pi^*(y\mid x)}{\pi_{\text{ref}}(y\mid x)} + \beta \log Z(x).
    \end{aligned}
\end{equation}

For $(r^*(x_1, y_w) - r^*(x_1, y_l))$ and $((r^*(x_2, y_w) - r^*(x_2, y_w))$ in Eqn.~\ref{eqa:appx_b_inter_level_RM_bt_derive}, we can substitute Eqn.~\ref{eqa:appx_B_DPO_rstar} and simplify:

\begin{equation}
    \begin{aligned}
    \label{eqa:appx_B_r-r}
        r^*(x_1, y_w) &- r^*(x_1, y_l) \\ &= \beta \log \frac{\pi^*(y_w\mid x_1)}{\pi_{\text{ref}}(y_w \mid x_1)} + \beta \log Z(x_1) - \beta \log \frac{\pi^*(y_l\mid x_1)}{\pi_{\text{ref}}(y_l \mid x_1)} - \beta \log Z(x_1) \\ &= \beta \log \frac{\pi^*(y_w\mid x_1)}{\pi_{\text{ref}}(y_w \mid x_1)} - \beta \log \frac{\pi^*(y_l\mid x_1)}{\pi_{\text{ref}}(y_l \mid x_1)} \\
        r^*(x_2, y_w) &- r^*(x_2, y_l) \\ &= \beta \log \frac{\pi^*(y_w\mid x_2)}{\pi_{\text{ref}}(y_w \mid x_2)} + \beta \log Z(x_2) - \beta \log \frac{\pi^*(y_l\mid x_2)}{\pi_{\text{ref}}(y_l \mid x_2)} - \beta \log Z(x_2) \\ &= \beta \log \frac{\pi^*(y_w\mid x_2)}{\pi_{\text{ref}}(y_w \mid x_2)} - \beta \log \frac{\pi^*(y_l\mid x_2)}{\pi_{\text{ref}}(y_l \mid x_2)}
    \end{aligned}
\end{equation}

Substituting the simplified result from Eqn.~\ref{eqa:appx_B_r-r} into Eqn.~\ref{eqa:appx_b_inter_level_RM_bt_derive}, we obtain:

\begin{equation}
    \begin{aligned}
    \label{eqa:appx_B_inter_level_DPO_bt}
        &p^*((x_1, y_w, y_l) \succ (x_2, y_w, y_l)) = \\ & \sigma \left( \left(\beta \log\frac{\pi^*(y_w \mid x_1)}{\pi_\text{ref}(y_w \mid x_1)} - \beta \log\frac{\pi^*(y_l \mid x_1)}{\pi_\text{ref}(y_l \mid x_1)}\right)-\left(\beta \log\frac{\pi^*(y_w \mid x_2)}{\pi_\text{ref}(y_w \mid x_2)} - \beta \log\frac{\pi^*(y_l \mid x_2)}{\pi_\text{ref}(y_l \mid x_2)}\right)\right)
    \end{aligned}
\end{equation}

Finally, we derive the loss function for inter-sample level DPO for a parameterized policy $\pi_\theta$ using the inter-sample level preference dataset $\mathcal{D}_\textit{inter} = \{(x_w, x, y_w, y_l)\} \cup \{(x, x_l, y_w, y_l)\}$ by applying maximum likelihood estimation:

\begin{equation}
    \label{eqa:appx_B_dpo_loss_inter}
    \begin{alignedat}{2}
        & \mathcal{L}_{\textit{DPO}}^{\textit{inter}}(\pi_\theta; \pi_\text{ref}) = \\
        & -\underset{(x_w, x, y_w, y_l) \sim \mathcal{D}_{\textit{inter}}}{\mathbb{E}}[\log\sigma( \,
        && (\beta \log\frac{\pi_\theta(y_w \mid x_w)}{\pi_\text{ref}(y_w \mid x_w)} - \beta \log\frac{\pi_\theta(y_l \mid x_w)}{\pi_\text{ref}(y_l \mid x_w)}) \\
        &&& - (\beta \log\frac{\pi_\theta(y_w \mid x)}{\pi_\text{ref}(y_w \mid x)} - \beta \log\frac{\pi_\theta(y_l \mid x)}{\pi_\text{ref}(y_l \mid x)}))] \\
        & -\underset{(x, x_l, y_w, y_l) \sim \mathcal{D}_{\textit{inter}}}{\mathbb{E}} [\log\sigma(\,
        && (\beta \log\frac{\pi_\theta(y_w \mid x)}{\pi_\text{ref}(y_w \mid x)} - \beta \log\frac{\pi_\theta(y_l \mid x)}{\pi_\text{ref}(y_l \mid x)}) \\
        &&& - (\beta \log\frac{\pi_\theta(y_w \mid x_l)}{\pi_\text{ref}(y_w \mid x_l)} - \beta \log\frac{\pi_\theta(y_l \mid x_l)}{\pi_\text{ref}(y_l \mid x_l)}))].
    \end{alignedat}
\end{equation}

\section{Implementation Details of M-LIFT}
\label{apx:M-LIFT_implement}

Original LIFT enhances existing preference data pairs by adding length instructions to prompts and reconstructing preference based on these added instructions to generate training dataset. Next, we introduce M-LIFT, which applies LIFT to multi-instruction following tasks. Specifically, given a preference dataset $\mathcal{D} = \{(x, y_w, y_l)\}$, we construct the dataset required for the M-LIFT method as follows:

First, for each preference data $(x, y_w, y_l)$ in $\mathcal{D}$, we randomly select functions $G_1 = \{g_{i_1}, \dots, g_{i_k}\}$ from function set $G_0 = \{g_1, g_2, ..., g_n\}$, provide conditions $\{(y_w, True/False), (y_l, True/False)\}$ for each $g_{i_k}$, and combine all returned instructions with $x$ to generate a multi-instruction prompt $x'$, forming $(x', y_w, y_l)$. We generate two categories of data that respectively satisfy the following two situations: 
1) $y_w$ satisfies all instructions and $y_l$ fails to satisfy at least one of them, with the generated multi-instruction prompt denoted as $x^1$. Thus, we obtain a new multi-instruction preference data $(x^1, y_w, y_l)$.
2) $y_l$ satisfies all instructions and $y_w$ fails to satisfy at least one of them, with the generated multi-instruction prompt denoted as $x^2$. Since $y_l$ performs better than $y_w$ in instruction following, we reverse the preference between $y_w$ and $y_l$, obtaining another new multi-instruction preference data $(x^2, y_l, y_w)$.

In this way, we obtain the dataset $\mathcal{D}_\textit{M-LIFT} = \{(x, y_w, y_l)\} \cup \{(x^1, y_w, y_l)\} \cup \{(x^2, y_l, y_w)\}$ required for M-LIFT training. Subsequently, we train the model using RM and DPO with $\mathcal{D}_\textit{M-LIFT}$.

\section{The Details of Dataset Generation for Training and Evaluation}
\label{apx:reward_model_eval}

In this section, we introduce the generation process for datasets $\mathcal{D}$, $\mathcal{D}_\textit{BT}$, $\mathcal{D}_\textit{eval}$, and $\mathcal{D}_\textit{eval}^\textit{if}$.

We extract the initial turn from each English conversation in the OpenAssistant dataset~\citep{köpf2023openassistantconversationsdemocratizing}. According to the human annotated preference information, responses with rank $0$ are labeled as the preferred choice ($y_w$), whereas those with rank $1$ are labeled as the rejected alternative ($y_l$). This process yields the original preference dataset $\mathcal{D} = \{(x, y_w, y_l)\}$.

We partition $\mathcal{D}$ into a training dataset $\mathcal{D}_\textit{BT}$ (comprising 90\% of $\mathcal{D}$) and an evaluation dataset $\mathcal{D}_\textit{eval}$ (comprising 10\% of $\mathcal{D}$). Here, $\mathcal{D}_\textit{BT}$ serves as the original preference dataset required for the \ourm{} framework, while $\mathcal{D}_\textit{eval}$ functions as evaluation dataset for assessing semantic quality of the trained model during the RM phase.

Additionally, we generate a multi-instruction evaluation dataset $\mathcal{D}_\textit{eval}^\textit{if}$ based on $\mathcal{D}_\textit{eval}$. Specifically, similar to section \ref{sec:3.2}, for each preference data $(x, y_w, y_l)$ in $\mathcal{D}_\textit{eval}$, we generate preference pairs for both $y_w$ and $y_l$, denoted as $(x_w^1, x_l^1, y_w)$ and $(x_w^2, x_l^2, y_l)$, where $x_w^1$ is superior to $x_l^1$, and $x_w^2$ is superior to $x_l^2$. This process yields the multi-instruction following evaluation dataset $\mathcal{D}_\textit{eval}^\textit{if} = \{(x_w^1, x_l^1, y_w)\} \cup \{(x_w^2, x_l^2, y_l)\}$.

\section{Broader Impact}
\label{apx:impact}

The goal of this work is to enhance the multi-instruction following capabilities of LLMs. When applied correctly, our advancements enable users to acquire desired information more rapidly, accurately and effectively, while simultaneously reducing the probability of users being misled by erroneous information resulting from improper instruction adherence. However, if the advancements are exploited for illegal activities, they could exert greater societal impact, thus necessitating strengthened regulatory oversight of LLMs.


\end{document}